\pdfoutput=1

\documentclass[11pt]{article}

\usepackage{ACL2023}
\usepackage{graphicx}
\usepackage{booktabs}
\usepackage{tabularx}
\usepackage{times}
\usepackage{latexsym}
\usepackage[T1]{fontenc}
\usepackage[utf8]{inputenc}
\usepackage{microtype}
\usepackage{amsmath}
\usepackage{hyperref}
\usepackage[hidecomments]{ufdatastudio}
\usepackage{multirow}
\usepackage{cleveref}

\title{BM25-Augmented Many-Shot Translation for Low-Resource North-Eastern Indian Languages}

\author{
  Aashish Dhawan$^{*}$ \\
  University of Florida \\
  \texttt{aashish.dhawan@ufl.edu}
  \And 
  Christopher Driggers-Ellis$^{*}$ \\
  University of Florida \\
  \texttt{driggersellis.cw@ufl.edu}
  \AND
  Dzmitry Kasinets$^{*}$ \\
  University of Florida \\
  \texttt{dkasinets@ufl.edu}
  \And
  Christan Grant \\
  University of Florida \\
  \texttt{christan@ufl.edu}
  \And
  Daisy Zhe Wang \\
  University of Florida \\
  \texttt{daisyw@cise.ufl.edu}
}

\begin{document}
\maketitle

\newcommand{\grn}{Guaraní }
\newcommand{\bzd}{Bribri }
\newcommand{\yua}{Yucatec Maya }
\newcommand{\nlv}{Orizaba Nahuatl }
\newcommand{\hch}{Wixárika }

\newcommand{\as}{Assamese }
\newcommand{\lus}{Mizo }
\newcommand{\kha}{Khasi }
\newcommand{\mni}{Manipuri }
\newcommand{\mtei}{Meitei }
\newcommand{\njz}{Nyishi }

\newcommand{\bodo}{Bodo }
\newcommand{\trp}{Kokborok }
\newcommand{\mjw}{Karbi }
\newcommand{\nag}{Nagamese }
\newcommand{\tgj}{Tagin }

\renewcommand{\thefootnote}{\fnsymbol{footnote}}
\footnotetext[1]{$^{*}$Equal contribution.}
\renewcommand{\thefootnote}{\arabic{footnote}}

\begin{abstract}
This paper describes the University of Florida gators submission to the WMT26 Low-Resource Indic Language Translation shared task. We adapt the retrieval-augmented many-shot translation pipeline from our AmericasNLP 2026 system~\cite{dhawan-etal-2026-retrieval,bui-etal-2026-findings} to translate between English and eleven North-Eastern Indian languages in both directions. At inference time, BM25~\cite{robertson2009probabilistic} retrieves the most similar parallel examples from a language-specific training bank, and Gemini 2.5 Flash~\cite{comanici2025gemini25} translates the input conditioned on these examples. No model fine-tuning is involved. Training banks combine official WMT26 data with publicly available corpora such as Samanantar~\cite{ramesh2022samanantar} and prior WMT shared task releases. A grid search over retrieval count~$r$ and development exemplar count~$d$ across all 22 language--direction pairs selects the best configuration for each submission. 
Find our code on \href{https://github.com/dhawan98/Gators_wmt26/tree/main}{Github}.
\end{abstract}

\section{Introduction}\label{sec:introduction}

The WMT26 Low-Resource Indic Language Translation shared task~\cite{parkay-etal-2026-findings-indic} covers eleven North-Eastern Indian languages that span four language families: Assamese (Indo-Aryan), Khasi (Austroasiatic), eight Tibeto-Burman languages (Mizo, Manipuri in both Bengali and Meitei Mayek scripts, Bodo, Kokborok, Karbi, Nyishi, and Tagin), and Nagamese, an English-based creole. Several of these --- particularly Nyishi~\cite{kakum2023neural}, Karbi, and Tagin --- have almost no prior MT work.

The system we describe here emerges from two earlier projects. In our LoResMT 2026 paper~\cite{dhawan-etal-2026-improving}, we found that augmenting mBART with NLLB-200-generated synthetic data improved ChrF++ for low-resource Indigenous American languages, but the gains were modest and domain-sensitive. We then moved to a retrieval-augmented approach for the AmericasNLP 2026 captioning task~\cite{dhawan-etal-2026-retrieval,bui-etal-2026-findings}: instead of fine-tuning, we used BM25 to pull relevant parallel examples into a long Gemini 2.5 Flash~\cite{comanici2025gemini25} prompt and let the model translate in context. That system won the overall shared task competition across five Indigenous American languages, and the gap over fine-tuned mBART was large enough that we wanted to test whether the approach would transfer to a different language family and task domain.

For WMT26, the adaptation is straightforward. The AmericasNLP pipeline had two stages --- VLM image captioning in Spanish, then retrieval-augmented Spanish-to-LoRes translation. Since WMT input is already English text, we drop the vision stage and feed source sentences straight into the BM25+Gemini pipeline. The rest carries over: per-language system prompts, the $r, d$ hyperparameter sweep, and the same scoring infrastructure.

\begin{figure*}[t]
    \centering
    \includegraphics[width=\textwidth]{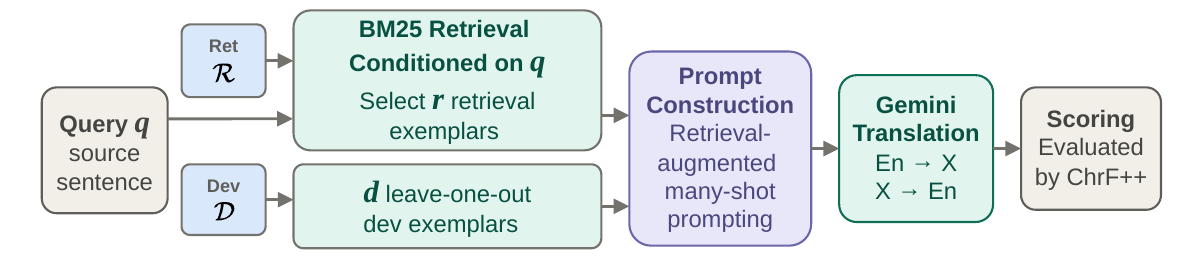}
    \caption{Overview of the \textbf{\texttt{gators}} pipeline.}
    \label{fig:wmt26-pipeline}
\end{figure*}

\section{Background}\label{sec:background}

The WMT Low-Resource Indic shared task has run annually since 2023~\cite{pal-etal-2023-findings-indic}, growing from four language pairs to eleven in 2026 \cite{parkay-etal-2026-findings-indic}. The task splits languages into two categories: Category~1 (moderate data) includes Assamese, Mizo, Khasi, Manipuri in Bengali script, Manipuri in Meitei Mayek, and Nyishi; Category~2 (very limited data) includes Bodo, Kokborok, Karbi, Nagamese, and Tagin. 
The Indic shared task organizers produce no official ranking of submissions, nor is a single winner declared, but they do report numerous metrics including BLEU \cite{papineni2002bleu}, ChrF++ \cite{popovic-2017-chrfpp},  COMET \cite{rei2020comet}, and others.

Previous editions~\cite{pakray-etal-2024-findings-indic,pakray-etal-2025-findings-indic} have been dominated by fine-tuned multilingual models --- IndicTrans2~\cite{gala2023indictrans2}, mBART~\cite{liu2020mbart}, and NLLB-200~\cite{costa2022nllb} --- combined with back-translation~\cite{sennrich2016improving} or transfer from related higher-resource Indic languages (see~\citealt{ranathunga2023neural} for a survey). 

Our submission takes a different route. The translation budget goes entirely to Gemini API calls, with BM25 handling example selection at inference time. This follows work on retrieval-augmented in-context translation~\cite{vilar2023prompting,agrawal2023context} showing LLMs can match or exceed fine-tuned baselines given well-chosen parallel demonstrations~\cite{brown2020language}. Related work from our lab on multimodal low-resource data generation~\cite{xiao-etal-2025-text} and cross-script parallel data~\cite{driggersellis2025multiscript30k} also informs the data preparation.

\section{System Description}\label{sec:system}

The overview of the pipeline is given in \Cref{fig:wmt26-pipeline}.
In this section, we will describe the phases of this pipeline from start to finish.

\subsection{Data Preparation}\label{subsec:data}

The retrieval bank $\mathcal{R}$ is built from WMT26 training data and
augmented with WMT23 pairs and publicly available parallel corpora. We retain
21 Manipuri and 3,554 Mizo pairs unique to WMT23%
~\cite{pal-etal-2023-findings-indic}.

\paragraph{Category 1 Languages}
For Assamese, we add 142,080 pairs from Samanantar%
~\cite{ramesh2022samanantar}, 1,359 from Endangered Recipes Translated
500~\cite{karya2026endangeredrecipestranslated500,karya2025elr1000,
joshi2025elr1000}, and 2,009 from IndicGenBench FLORES%
~\cite{singh2024indicgenbench} to the 54,000 WMT26 pairs. Khasi receives
1,048 Endangered Recipes pairs in addition to 26,000 WMT26 pairs, while
Manipuri receives 2,009 FLORES pairs and 21 WMT23 pairs in addition to
23,687 WMT26 pairs.

We add no WMT-external data for the remaining Category 1 languages. 
Meitei-Mayek has 17,001
WMT26 pairs. Nyishi has 60,000 WMT26 pairs, and Mizo has 53,543 pairs
across WMT26 and WMT23---49,989 and 3,554 pairs, respectively. We
focus our data-collection efforts on the more resource-constrained
Category 2 languages, although additional parallel data may exist for
these three languages. 
Remaining paragraphs in this subsection will detail data augmentation for each Category 2 language separately.

\paragraph{Bodo}
We combine 15,214 WMT26 pairs with 134,931 pairs from the English--Bodo
Parallel Dataset~\cite{panjiyar_english_bodo}, 1,497 Endangered Recipes
pairs, and 2,009 FLORES pairs to produce the largest corpus of WMT and external data.

\paragraph{Karbi}
We add 382 pairs mined from \textit{A Grammar of Karbi}%
~\cite{konnerth2014grammarofkarbi} and 1,144 from \textit{Karbi Texts}%
~\cite{konnerth2018karbitexts} to the 1,000 WMT26 pairs.

\paragraph{Kokborok, Nagamese, and Tagin}
The corresponding WMT26 sets contain 2,269, 1,998, and 5,344 pairs. We add
6,078 Kokborok pairs from the Kokborok Digitalisation Project%
~\cite{caswell2025smol,jones2023gatitos}, 8,771 English--Nagamese pairs
from FNPC-25~\cite{maheshwari2025fnpc}, and 854 English--Tagin pairs from
GinLish v0.1.0~\cite{dugi2024ginlish}.

All sources are merged using SHA-256 source-side deduplication, Unicode
normalization (NFC for Bengali and Devanagari and NFKC for Latin scripts),
and length-ratio filtering. For noisier external sources, we optionally run LaBSE~\cite{feng2022labse} cross-lingual similarity filtering.

\subsection{In-Domain Data Filtering}\label{subsec:retrieval}
We restrict retrieval to in-domain data because sentence pairs from unrelated domains have little lexical or semantic overlap with the WMT26 inputs. As a result, BM25 assigns them low relevance scores, making them unlikely to be selected as useful in-context examples. We use filtered, in-domain data in both our primary and contrastive systems. The final primary corpus sizes \textbf{Train} and number of \textbf{Pairs}, including contrastive data, appear in \Cref{tab:in-domain-data}.

\begin{table}[!t]
\centering
\small
\setlength{\tabcolsep}{5pt}
\begin{tabular}{@{}cll|rrr@{}}
\toprule
\textbf{Cat.} &
\textbf{Language} &
\textbf{ISO} &
\textbf{Train} &
\textbf{Aug.} &
\textbf{Pairs} \\
\midrule
\multirow{6}{*}{1}
& \as            & As          & 52,440 & 3.3$\times$ & 171,794 \\
& \lus           & Lus         & 49,293 & 1.0$\times$ &  50,620 \\
& \kha           & Kha         & 25,495 & 0.9$\times$ &  22,847 \\
& \mni           & Mni         & 22,539 & 1.0$\times$ &  22,544 \\
& \mtei & Mni\_Mtei & 16,312 & 1.0$\times$ &  16,312$\dagger$ \\
& \njz           & Njz         & 49,613 & 1.0$\times$ &  49,793 \\
\midrule
\multirow{5}{*}{2}
& \bodo          & Bodo        & 12,268 & 9.3$\times$ & 114,410 \\
& \trp           & Trp         &  2,040 & 1.0$\times$ &   2,102 \\
& \mjw           & Mjw         &    864 & 1.0$\times$ &     899 \\
& \nag           & Nag         &  1,798 & 5.7$\times$ &  10,169 \\
& \tgj           & Tgj         &  4,703 & 0.9$\times$ &   4,421 \\
\bottomrule
\end{tabular}
\caption{Sizes of the in-domain training data before and after augmentation. The primary system uses the \textbf{Train} data, while the contrastive system uses the \textbf{Pairs} data. The augmentation ratio is computed as the number of augmented sentence pairs divided by the number of base pairs. $\dagger$ \textbf{Pairs} recalculated from \textbf{Train}. They are equal because we use no external data for Mni\_Mtei. 
}
\label{tab:in-domain-data}
\end{table}


\subsection{BM25 Retrieval}\label{subsec:retrieval}

The source side of $\mathcal{R}$ is indexed with BM25 Okapi~\cite{robertson2009probabilistic} ($k_1 = 1.5$, $b = 0.75$). Given a source sentence $q$, the system retrieves the top-$r$ training pairs $\mathcal{R}_r(q) \subseteq \mathcal{R} $, skipping exact matches on the query and deduplicating by 60-character normalized prefix to prevent overlapping corpora (e.g., Samanantar vs. PMIndia) from flooding the prompt with near-identical examples.

Getting the index direction right matters. For En$\to$X, the index is built on the English sentences. For X$\to$En, the query is in the Indic language, so the index must be built on the Indic-language side. We had a bug in the AmericasNLP system~\cite{dhawan-etal-2026-retrieval} where Spanish (Es) was always indexed regardless of direction, but it was overlooked because X$\to$Es translation was not necessary. We find and resolve the issue in this shared task.

Configurations additionally include $d$ development exemplars $\mathcal{D}_d \subseteq \mathcal{D}$, selected by word-overlap ranking with leave-one-out exclusion. The prompt context becomes $P(q) = \mathcal{D}_d \cup \mathcal{R}_r(q)$. We conduct a grid search over $r \in \{0, 20, 40, 80\}$ and $d \in \{0, 10, 20, 49\}$ for each language and direction --- a 4$\times$4 grid of 16 configurations per language--direction pair, evaluated on 50 held-out development sentences per cell. The grid is complete across all 22 language--direction pairs.

\subsection{Translation}\label{subsec:translation}

Prompts go to Gemini 2.5 Flash~\cite{comanici2025gemini25} via the \texttt{google-genai} SDK (temperature~0, thinking disabled, max output 512 tokens). We use a 2-key API pool with round-robin and quota failover to manage rate limits. Each of the eleven languages has its own system prompt encoding script, word order, and morphological type --- the Khasi prompt warns about VOS order and diacritics, the Bodo prompt states that despite Devanagari script the language is Tibeto-Burman and not Hindi. 
We elaborate these prompting considerations in our appendix. 

Post-processing strips model-generated prefixes, applies Unicode normalization, and for non-Latin scripts validates that output characters fall within the expected Unicode block. The validation is especially important for Meitei Mayek (U+ABC0--U+ABFF), where Gemini occasionally falls back to Bengali. A negligible fraction of output lines (4 of 21,599; 0.018\%) are empty due to Gemini safety filters on specific political sentences that temperature-perturbed retry does not resolve.

\subsection{Relation to the AmericasNLP System}

As shown in Figure~\ref{fig:wmt26-pipeline}, the main architectural change from the AmericasNLP pipeline~\cite{dhawan-etal-2026-retrieval} is the removal of the vision-captioning stage. We also add bidirectional translation support through the BM25 index fix described, and merge multiple external corpora into retrieval banks.

\section{Languages}\label{sec:languages}

Most language tracks in the shared task---eight of eleven---represent Tibeto-Burman languages, while the three non-Tibeto-Burman languages are Assamese, Khasi, and Nagamese~\cite{post2017tibeto}. The languages are predominantly Subject--Object--Verb (SOV), consistent with the clause-final pattern common across the Tibeto-Burman and local Indo-Aryan languages of Northeast India. Khasi, an Austroasiatic language, is the principal exception, with basic Subject--Verb--Object (SVO) order~\cite{post2017tibeto,sharma1999comparison}. Assamese is an Indo-Aryan language, while Nagamese is an Assamese-lexified creole spoken across Nagaland~\cite{post2017tibeto,maiti2025naganlp}. Manipuri appears in two task tracks, one written in Bengali script and the other in Meitei Mayek~\cite{singh2012bidirectional}. These tracks represent two scripts of the same language rather than genetically distinct languages; we refer to the Meitei Mayek track as Meitei elsewhere in the paper. Bodo belongs to the Tibeto-Burman family but is written in the Devanagari script~\cite{post2017tibeto,narzary2022generating}. In our experiments, Gemini occasionally produced Hindi-like Bodo output, likely because Bodo shares the Devanagari script with Hindi and
Nepali~\cite{narzary2022generating}. Explicitly identifying Bodo in the
system prompt removed this behavior.

\section{Experimental Setup}\label{sec:experiments}

Experiments run on the University of Florida HiPerGator cluster. The translation pipeline is CPU-only (it makes API calls to Gemini) and runs on the \texttt{hpg-default} partition. The environment is a Python venv (not conda). GPU nodes (\texttt{hpg-b200}) and mamba (conda) environments are used only for NLLB-200 data generation Evaluation uses SacreBLEU to measure ChrF~\cite{popovic-2015-chrf} and ChrF++~\cite{popovic-2017-chrfpp}.

Since no official WMT26 development set was released early on, we hold the last 500 pairs or fewer from each language's \textbf{Train} data as the set of possible development exemplars $\mathcal{D}$ ($|\mathcal{D}| \le 500$) before adding external data to obtain \textbf{Pairs}. We record $|\mathcal{D}|$ for each language in \Cref{tab:card-d}. 
The pilot grid search evaluates each $(r, d)$ cell on $n = 50$ sentences from this split.
We submit the configuration that achieves the greatest ChrF score.

\begin{table}[t]
    \centering
    \small
    \begin{tabular}{cll|r}
    \toprule
\textbf{Cat.} &
\textbf{Language} &
\textbf{ISO} &
\multicolumn{1}{c}{\textbf{$|\mathcal{D}|$}} \\
\midrule
\multirow{6}{*}{1}
& \as            & As          &  500\\
& \lus           & Lus         &  500 \\
& \kha           & Kha         &  500 \\
& \mni           & Mni         &  500\\
& \mtei & Mni\_Mtei &  500 \\
& \njz           & Njz         &  500\\
\midrule
\multirow{5}{*}{2}
& \bodo          & Bodo        &  500 \\
& \trp           & Trp         &    500 \\
& \mjw           & Mjw         &        96 \\
& \nag           & Nag         &   199 \\
& \tgj           & Tgj         &    226 \\
\bottomrule
\end{tabular}
    \caption{$|\mathcal{D}|$ for each language in the shared task. Note that $|\mathcal{D}| = 500$ except for some Cat. 2 languagues.}
    \label{tab:card-d}
\end{table}

\section{Results}\label{sec:results}

\begin{table*}[t]
\centering
\small
\begin{tabular}{cllrr|rrrr|rrrr}
\toprule
\multicolumn{5}{c}{Invariant} &
\multicolumn{4}{|c}{\texttt{dev En$\to$X}} & \multicolumn{4}{|c}{\texttt{dev X$\to$En}} \\
\midrule
Cat. & Language & ISO & Aug. & Pairs & \multicolumn{1}{r}{$r$} & \multicolumn{1}{r}{$d$} & \multicolumn{1}{r} {\texttt{ChrF}} & \multicolumn{1}{r} {\texttt{ChrF++}} & \multicolumn{1}{|r}{$r$} & \multicolumn{1}{r}{$d$} & \multicolumn{1}{r} {\texttt{ChrF}} & \multicolumn{1}{r} {\texttt{ChrF++}} \\
\midrule
\multirow{6}{*}{1} 
&\as & As & 3.3$\times$ & 171,794 & 80 & 10 & 52.95 & 49.81 & 40 & 49 & 68.80 & 67.98 \\
&\lus & Lus & 1.0$\times$ & 50,620 & 80 & 20 & 53.40 & 51.90 & 40 & 20 & 63.53 & 62.53 \\
&\kha & Kha & 0.9$\times$ & 22,847 & 80 & 49 & 56.44 & 55.71 & 80 & 49 & 75.64 & 75.09 \\
&\mni & Mni & 1.0$\times$ & 22,544 & 80 & 49 & 66.72 & 62.34 & 20 & 49 & 69.97 & 67.86 \\
&\mtei & Mni\_Mtei & 1.0$\times$ & 16,312 & 80 & 10 & 41.22 & 36.86 & 40 & 10 & 55.20 & 52.85 \\
&\njz  & Njz & 1.0$\times$ & 49,793 & 80 & 20 & 44.98 & 40.58 & 80& 20& 58.76 & 57.96 \\
\midrule
\multirow{5}{*}{2} 
&\bodo & Bodo & 9.3$\times$ & 114,410 & 80 & 49 & 76.37 & 72.14 & 20 & 10 & 79.62 & 78.38 \\
&\trp & Trp & 1.0$\times$ & 2,102 & 80 & 49 & 54.69 & 51.61 & 80 & 49 & 86.90 & 86.26 \\
&\mjw & Mjw & 1.0$\times$ & 899 & 40 & 49 & 49.19 & 44.84 & 40 & 49 & 62.71 & 60.82 \\
&\nag & Nag & 5.7$\times$ & 10,169 & 80 & 20 & 55.70 & 52.59 & 80 & 20 & 66.59 & 64.69 \\
&\tgj & Tgj & 0.9$\times$ & 4,421 & 80 & 10 & 46.02 & 43.33 & 40 & 20 & 47.05 & 45.77 \\
\bottomrule
\end{tabular}
\caption{Development contrastive submission results for each combination of language and translation direction. 
Two optimal $r,d$ pairs are selected for each language from our search grid, one per translation direction. 
\textbf{Augmentation (Aug.)} gives a rough multiplier on the original size of the development set that our contrastive data augmentation provides. 
\textbf{Pairs} is the total number of contrastive retrieval exemplars $|\mathcal{R}|$ after augmentation.}
\label{tab:main_results_dev}
\end{table*}

\begin{table*}[t]
\centering
\small
\begin{tabular}{cllrr|rrrr|rrrr}
\toprule
\multicolumn{5}{c}{Invariant} &
\multicolumn{4}{|c}{\texttt{test En$\to$X}} & \multicolumn{4}{|c}{\texttt{test X$\to$En}} \\
\midrule
Cat. & Language & ISO & Aug. & Pairs & \multicolumn{1}{r}{$r$} & \multicolumn{1}{r}{$d$} & \multicolumn{1}{r} {\texttt{BLEU}} & \multicolumn{1}{r|}{\texttt{ChrF++}} & \multicolumn{1}{|r}{$r$} & \multicolumn{1}{r}{$d$} & \multicolumn{1}{r} {\texttt{BLEU}} & \multicolumn{1}{r}{\texttt{ChrF++}}\\
\midrule
\multirow{6}{*}{1} 
&\as & As & 3.3$\times$ & 171,794 & 80 & 10 & \textbf{20.77} & \textbf{52.81} & 40 & 49 & \textbf{27.27} & \textbf{57.35} \\
&\lus & Lus & 1.0$\times$ & 50,620  & 80 & 20 & 21.60 & 50.69 & 40 & 20 & 33.28 & \textbf{63.36}\\
&\kha & Kha & 0.9$\times$ & 22,847 & 80 & 49 & 29.14 & 53.16 & 80 & 49 & 25.76 & \textbf{52.92} \\
&\mni & Mni & 1.0$\times$ & 22,544 & 80 & 49 & 8.15& \textbf{39.89} & 20 & 49 & \textbf{45.93} & \textbf{72.37} \\
&\mtei & Mni\_Mtei & 1.0$\times$ & 16,312 & 80 & 10 & \textbf{16.62} & \textbf{52.85} & 40 & 10 & \textbf{38.13} & \textbf{66.16} \\
&\njz & Njz & 1.0$\times$ & 49,793  & 80 & 20 & -- & --& 80& 20& --& -- \\
\midrule
\multirow{5}{*}{2} 
&\bodo & Bodo & 9.3$\times$ & 114,410 & 80 & 49 & 14.61 & 48.21 & 20 & 10 & 34.10 & 62.10 \\
&\trp & Trp & 1.0$\times$ & 2,102 & 80 & 49 & 6.02 & 33.70 & 80 & 49 & 16.67 & 41.58 \\
&\mjw & Mjw & 1.0$\times$ & 899 & 40 & 49 & \textbf{5.74} & \textbf{33.85} & 40 & 49 & -- & -- \\
&\nag & Nag & 5.7$\times$ & 10,169 & 80 & 20 & \textbf{16.62} & \textbf{54.14} & 80 & 20 & \textbf{19.48} & \textbf{44.66} \\
&\tgj & Tgj & 0.9$\times$ & 4,421 & 80 & 10 & 3.31 & 24.13 & 40 & 20 & \textbf{4.01} & 17.73 \\
\bottomrule
\end{tabular}
\caption{Test contrastive submission results for each combination language and translation direction \cite{parkay-etal-2026-findings-indic}. 
Two optimal $r,d$ pairs are selected for each language from our search grid, one per translation direction. 
\textbf{Augmentation Ratio (Aug.)} gives a rough multiplier on the original size of the development set that our contrastive data augmentation provides. 
\textbf{Pairs} is the total number of contrastive retrieval exemplars $|\mathcal{R}|$ after augmentation.
Entries in \textbf{bold} indicate that our submission performs best in the shared task for that metric.}
\label{tab:main_results_test}
\end{table*}




In this section, we introduce the results of our method for translation of Category 1 and Category 2 languages in both the En$\to$X and X$\to$En translation tasks.
\Cref{tab:main_results_dev} lists the performance metrics for our contrastive approach for each language pair and translation direction, evaluated with the augmented development datasets prior to submission. 
We measure performance using the ChrF and ChrF++ metrics.
The former determined what $r,d$ pairs feature in our submissions, and we include ChrF++ for direct comparison of dev and test performance.

\Cref{tab:main_results_test} lists contrastive performance for the languages and translation directions according to the BLEU and ChrF++ measurements published by the shared task organizers \cite{parkay-etal-2026-findings-indic}.
Other test-time metrics are available in the shared task findings paper \cite{parkay-etal-2026-findings-indic}.


\section{Ablations}\label{sec:ablations}

In this section, we look at alternative approaches to the proposed method.
We hope these ablations will guide future work by showing how to improve or not to improve on the current method and by deepening the discussion to follow about its strengths and weaknesses.

\subsection{Multi30k Domain Mismatch}

The clearest negative result concerns synthetic data augmentation, which we apply to positive effect in multiple previous MT experiments \cite{dhawan-etal-2026-improving, dhawan-etal-2026-retrieval}. 

We prepare but ultimately \emph{exclude} synthetic data that would have originally augmented three of the eleven languages in the present task. 
Following the MultiScript30k methodology~\cite{driggersellis2025multiscript30k}, we generate NLLB-200 translations of the Multi30k image captions~\cite{costa2022nllb, elliot2016multi30ken} into Assamese, Mizo, and \mni ($\sim$29K pairs each). 
This augmentation is the single largest driver of \grn performance in our AmericasNLP system.
Multi30k captions are somewhat domain-matched to the captioning task \cite{dhawan-etal-2026-retrieval} because they are image captions, and the queries at the MT stage of our submission are machine-generated Spanish captions of images.
Meanwhile, WMT26 test data are sports news and government proceedings --- ``PCB requests ICC review,'' ``India loses 0-8 in hockey''.  

We conclude that domain match contributes more strongly to performance than sheer volume in BM25-based RAG and exclude MultiScript30k augmentation from all final retrieval banks.

\begin{figure*}[t]
    \centering
    \includegraphics[width=\textwidth]{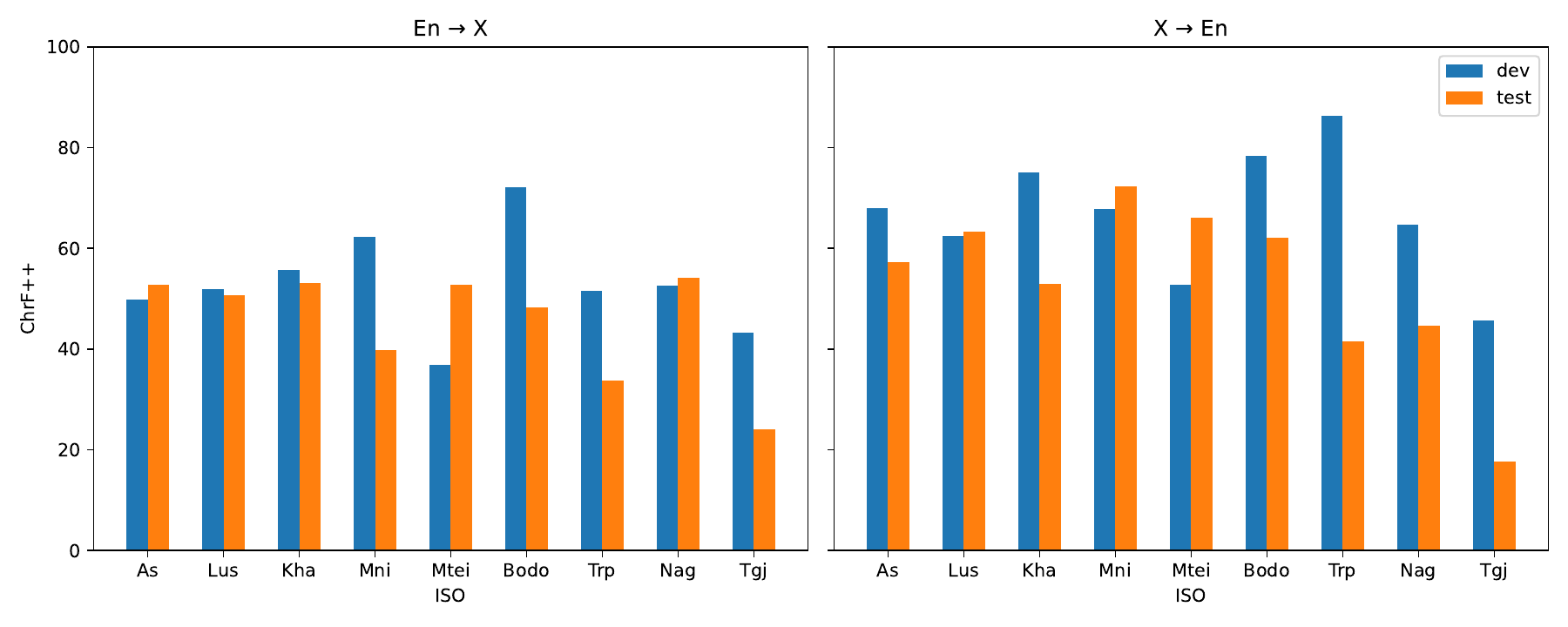}
    \caption{Dev-to-test performance changes in the contrastive system. \textbf{Left:} En$\to$X translation performance for dev (\textcolor{blue}{blue}) and test (\textcolor{brown}{orange}) data. \textbf{Right: } X$\to$En translation performance for dev (\textcolor{blue}{blue}) and test (\textcolor{brown}{orange}) data.}
    \label{fig:dev_v_test}
\end{figure*}

\begin{table*}[t]
\centering
\small
\begin{tabular}{cllrr|rrr|rrr}
\toprule
\multicolumn{5}{c}{Invariant} &
\multicolumn{3}{|c}{\texttt{En$\to$X}} & \multicolumn{3}{|c}{\texttt{X$\to$En}} \\
\midrule
Cat. & Language & ISO & Aug. & Pairs & \multicolumn{1}{r}{$r$} & \multicolumn{1}{r}{$d$} & \multicolumn{1}{r} {\texttt{ChrF++}} & \multicolumn{1}{|r}{$r$} & \multicolumn{1}{r}{$d$} & \multicolumn{1}{r} {\texttt{ChrF++}} \\
\midrule
\multirow{6}{*}{1} 
&\as & As & 3.3$\times$ & 171,794 & 80 & 20 & 49.78 &  80 & 49 & 79.00 \\
&\lus & Lus & 1.0$\times$ & 50,620 & 80 & 49 & 52.54 & 40 & 49 & 77.90\\
&\kha & Kha & 0.9$\times$ & 22,847 & \textit{80} & \textit{49} & 55.23 & 40 & 49 & 82.81 \\
&\mni & Mni & 1.0$\times$ & 22,544 & 80 & 20  & 62.96 & 0 & 49 & 88.88 \\
&\mtei & Mni\_Mtei & 1.0$\times$ & 16,312 & 80 & 49 & 36.39 & 20 & 49 & 55.08 \\
&\njz  & Njz & 1.0$\times$ & 49,793 & \textit{80} & \textit{20} & 39.83 & 0 & 49 & 72.93\\
\midrule
\multirow{5}{*}{2} 
&\bodo & Bodo & 9.3$\times$ & 114,410 & 80 & 20 &  73.36 & 0 & 49 & 81.87 \\
&\trp & Trp & 1.0$\times$ & 2,102 & \textit{80} & \textit{49} & 52.76 & 0 & 49 & 68.29 \\
&\mjw & Mjw & 1.0$\times$ & 899 & \textit{40} & \textit{49}  & 46.76 & 40 & 49 & 62.71 \\
&\nag & Nag & 5.7$\times$ & 10,169 & 40 & 10 & 53.87 & 0 & 20 & 53.52 \\
&\tgj & Tgj & 0.9$\times$ & 4,421 & 80 & 49 & 35.31 & 0 & 49 & 71.16\\
\bottomrule
\end{tabular}
\caption{Results of repeating our $r,d$ grid search on near-duplicate-filtered WMT26 data. 
\textbf{Augmentation (Aug.)} gives a rough multiplier on the original size of the development set that our contrastive data augmentation provides. 
\textbf{Pairs} is the total number of contrastive retrieval exemplars $|\mathcal{R}|$ after augmentation. 
Unchanged optimal $r,d$ configurations in \textit{italics}.}
\label{tab:second_results}
\end{table*}

\subsection{Key Observations from the Grid Search}


\paragraph{Script effect.} While they are treated as entirely separate languages by our method and the shared task documentation, we have already acknowledged that \mni is represented twice in the present task.
What we call \mni is \mni in Bengali script while the language called \mtei is \mni written in the \mtei Mayek script.
For the Bengali script, our submission achieves 62.34 dev ChrF++ in the En$\to$Mni direction, versus 36.86 dev ChrF++ for En$\to$Mni\_Mtei translation.
This trend repeats for translation in the X$\to$En direction but performance drops less steeply from 67.86 to 52.85 ChrF++.
This represents roughly 25- and 15-point gaps for the same spoken language.
We observe a similar trend for X$\to$En translation on test data that reverses during En$\to$X translation.

We conjecture that Gemini's pre-training exposure to \mni in Bengali script is likely far greater than \mni in Meitei Mayek (U+ABC0--U+ABFF), which appears much more scarcely in online text, but that cannot account for the inconsistencies we observe.
In any case, we caution that choice of orthography, where applicable, impacts MT performance using our RAG method.


\paragraph{Data Cleaning.} In addition to the original grid search described in \Cref{sec:experiments}, we perform the same procedure after the submission deadline using slightly different data and ranking.
We filter the WMT26 training data again, this time removing near duplicate candidate exemplars, and rank on the ChrF++ metric rather than ChrF.

The results appear in \Cref{tab:second_results}, and comparison with  \Cref{tab:main_results_dev} and \Cref{tab:main_results_test} reveals that optimal the $r,d$ configuration changes for 18 of 22 tasks.
In particular, the optimal $r$ falls to zero in seven out of eleven  X$\to$En translation tasks.
Conversely, nine out of eleven X$\to$En translation tasks now show $d = 49$.
Six have ($r = 0, d = 49$).
Nine out of eleven En$\to$X tasks still show $r = 80$.

These results for newly filtered data, although they still show performance for a small pilot of $n = 50$ samples, demonstrate a clear asymmetry with respect to translation direction that our original grid search does not show.
An optimum $r = 80$ for most languages in En$\to$X performance suggests that RAG does improve  performance while generating a low-resource language, as in these tasks and our original AmericasNLP shared task submission.
Two $r = 40$ configurations appear in \mjw and \nag translation, and these feature the smallest corpora, possibly indicating that retrieval strength is limited during extreme resource scarcity.
Meanwhile, $r = 0$ in X$\to$En tasks indicates that retrieval does not help when generating a high-resource language the LLM already knows well and when there are no samples identical or almost identical to the actual query to draw on.

\paragraph{Source-Side Retrieval Index.} For X$\to$En, building the BM25 index on the Indic language's side (rather than English in all settings) changes the optimal configuration for most languages. 
This is an essential improvement to our AmericasNLP 2026 submission \cite{dhawan-etal-2026-retrieval}.
Under the previous, source-invariant English BM25 index, the optimal $r,d$ configuration had $r = 0$ for As$\to$En and Kha$\to$En translation.
The implication is that no retrieval outperforms any retrieval based on the fixed English index for these languages.
When the BM25 index is corrected, the optimal $r$ increases to what we present in \Cref{tab:main_results_dev} and \Cref{tab:main_results_test}, indicating that retrieval does finally improve translation.

\section{Discussion}\label{sec:discussion}




For nearly every language and direction, we observe a some decay in performance from development to test set data.
We expect this decay given our previous results in the AmericasNLP shared task \cite{dhawan-etal-2026-retrieval}.
The greatest dev-test differences appear for the \bodo and \trp and for En$\to$Bodo translation and Trp$\to$En translation in particular.
We observe a difference of nearly 30 and 40 points, respectively, as we shift from the development to the test data.
Still, \bodo and \trp are outliers as compared to the other languages, and there are many cases where test performance is actually greater than dev performance according to the ChrF++ metric.
We build \Cref{fig:dev_v_test} in Matplotlib \cite{Hunter:2007} to demonstrate the performance gaps and exceptions to predicted performance decline for both En$\to$X (right) and X$\to$En (left) translation tasks.

For languages where the gap is thinner or flipped, this may a signal that the test dataset for the current shared task is much closer to the same domain as the development set than the datasets in AmericasNLP 2026's shared task, but a close examination of the test data or thorough documentation of it is necessary to confirm this hypothesis and it would not explain sizable performance gains as we move from development to test data.

Category 1 languages typically have many times more retrieval pairs than Category 2 languages, \bodo not withstanding.
Least-squares linear regression analysis with Scipy~\cite{2020SciPy-NMeth}, including Wald t-tests, reveals no significant ($p < 0.05$) correlation between \textbf{Pairs} and ChrF++ or BLEU performance, although we do uncover a weak but significant positive correlation ($r = 0.650$, $p < 0.05$, $m > 0$) between augmentation ratio and dev En$\to$X performance. 
These findings may be linked to our previous observation that total retrieval bank volume is less important than domain alignment.

The effect of $r$ and $d$ values in this shared task is well explored through our grid search.
For Category 2 languages, we observe that the gain from $d=0$ to $d=10$ exceeds the gain from $r=0$ to $r=80$, holding the other hyperparameter constant.
Linear regression analysis also discovers a significant positive correlation between dev $d$ values and En$\to$X ChrF++ ($r = 0.625$, $p < 0.05$, $m > 0$) but none between $r$ and En$\to$X preformance.
Curated development examplars therefore appear to be a stronger signal than BM25-retrieved training pairs when the corpus is small. 
$r = 80$ performs best for ten of the eleven languages in En$\to$X translation.
In particular, \bodo performance peaks at $(r = 80, d = 49)$ with a score of 76.37 ChrF and 72.14 ChrF++.
We investigate the search grids further and find that the performance gains as $r$ increases, holding $d$ constant, are somewhat flat for languages with less than 10,000 total exemplars, including  \mjw (899) and \tgj (4,421).
\mjw is the one language with $r < 80$ in our contrastive En$\to$X submissions.
We conclude larger retrieval corpora better reward higher retrieval counts $r$.

\section{Conclusion}\label{sec:conclusion}

We describe the gators submission to WMT26's Indic Shared Task: A training-free, retrieval-augmented system that uses BM25 and Gemini 2.5 Flash to translate between English and eleven Northeastern Indian languages. The approach applies transfer learning to our AmericasNLP 2026 system~\cite{dhawan-etal-2026-retrieval} but removes vision to isolate retrieval-augmented MT and builds retrieval banks with different parallel text.

Along with other ablations, we complete a grid search across all 22 language--direction pairs, and the submission covers every pair in both directions.
We show the importance of domain matching to the synthetic data augmentation we espouse in the previous shared task \cite{dhawan-etal-2026-retrieval} and fix a critical bug in our original BM-25 index construction.

While the previous shared task submission shows that retrieval-augmented MT can translate VLM-generated captions into low-resource languages, we now isolate the capabilities of retrieval-augmented MT and reproduce our earlier success.
Our submissions consistently outperform the competition by small yet measurable margins.
In 19 translation tasks, our contrastive submission achieves the best BLEU score for nine and the best ChrF++ score for eleven \cite{parkay-etal-2026-findings-indic}.

\section*{Limitations}\label{sec:limitations}

The system depends entirely on the Gemini 2.5 Flash API. 
During the pilot grid search, burst-level rate-limit 429 errors under exponential backoff consumed entire SLURM job windows, requiring careful scheduling and a 2-key API pool to complete all 22 directions' grid cells in time for submission. 
A small number of test sentences (4 of 21,599; 0.018\%) produced empty output due to Gemini's safety filters blocking specific political content.

Retrieval bank quality varies widely. Category~2 languages have limited parallel data, and the available external corpora are often out-of-domain. For instance, our only external corpus for \nag consists of Bible translations. 
As an ablation in \Cref{sec:ablations} shows, adding more data does not help when BM25 will not surface it.

The development split, consisting of the last 500 or fewer pairs of training data and with 50 sentences evaluated per grid cell, shares its domain with the training corpus; so dev metrics overstate performance. 
We see a similar effect in AmericasNLP~\cite{dhawan-etal-2026-retrieval}.
In particular, \grn dev scores roughly halved during evaluation with the official test set.
This time, \Cref{tab:main_results_dev} shows an inconsistent drop off for all languages except Bodo, but a performance drop usually exists.


\section*{Acknowledgments}
We thank Arnold and Lisa Goldberg for their financial support and UF Research Computing for HiPerGator access. 
We also  acknowledge the Gatorade, whose support provided the compute resources necessary for this research.

\bibliography{citations}
\bibliographystyle{acl_natbib}

\clearpage

\section*{Appendix}

This final supplementary section is dedicated to communicating fine-grained details of our submissions that are not in focus in the main body of our paper.
We summarize in \Cref{subsec:prompts} the prompt engineering that goes into calling Gemini 2.5 Flash during each experiment described and during testing for each language and translation direction.
\Cref{subsec:r-d} discusses the $r,d$ grid search we conduct over a pilot sample of the available WMT26 data to select retrieval hyperparameter configurations for our final submission. 

\subsection{Prompts}\label{subsec:prompts}

In \Cref{tab:prompts} and \Cref{tab:prompts2}, we describe the considerations taken at the prompting stage for each language and translation direction.
In general, we conditioned Gemini 2.5 Flash to generate the correct script and encoding during En$\to$X translation through our prompts.
We also inform Gemini about morphological considerations including word order, tonality, and type of agglutination in the Indic target language where applicable.
Because Gemini is asked to generate En during X$\to$En translation, we find it much less necessary to precondition generation in the high-resource language through extensive prompting.

\subsection{$r, d$ Search Grid}\label{subsec:r-d}

Similar to our AmericasNLP 2026 shared task submission \cite{dhawan-etal-2026-retrieval}, we define a search grid consisting of all $(r,d) \in \{0,20,40,80\} \times \{0,10,20,49\}$ and record ChrF and ChrF++ metrics for each tuple $(r,d)$ during an evaluation with a pilot set of $n = 50$ WMT26 samples.
We designate ChrF as the primary metric for the primary grid search, and our submission for each language and translation direction has the optimal $r,d$ hyperparameter configuration we determine through this grid search.
After the submission deadline, we repeat the procedure with deduplicated samples and rank according to the ChrF++ metric.

Inclusion in this appendix of a full search over the eleven shared task languages in both translation directions, a total of 22 translation tasks,  would imply tabulating  22 $\times$ 16 or 352 grid search cells in the following pages.
This is doubled to 704 cells if we include the second search's results.
Instead, we refer the reader to our Github page, where we host a \href{https://github.com/dhawan98/Gators_wmt26/blob/main/slurm_pilot_grid.sh}{bash script} that can recreate our grid search experiments using WMT26 data.

\begin{table*}
    \centering
    \small
    \begin{tabular}{ll|c|p{4.5cm}|c|p{4.5cm}}
        \toprule
        \multicolumn{2}{c}{Language} &
        \multicolumn{2}{|c}{\texttt{En$\to$X}} & \multicolumn{2}{|c}{\texttt{X$\to$En}} \\
        \midrule
        Lang. & ISO & URL & Considerations & URL & Considerations \\
        \midrule
        \as & As & \href{https://github.com/dhawan98/Gators_wmt26/blob/292f0922be4aea1aeedf9e2159f83dd52123c3af/translate_llm_manyshot_gemini.py#L112}{URL} & \begin{tabular}{l}
             Agglutinative Postpositions \\
             Bengali-Assamese Script \\ 
             Honorifics \\
             Match Example Register\\
             Match Example Style\\
             No Addtl. En,  Expl., Parens\\
             Output Only Single Line of As\\
             SOV Word Order \\ 
             Unicode NFC Encoding \\ Unique Assamese Characters \\
        \end{tabular}& \href{https://github.com/dhawan98/Gators_wmt26/blob/292f0922be4aea1aeedf9e2159f83dd52123c3af/translate_llm_manyshot_gemini.py#L126}{URL} & \begin{tabular}{l}
             Agglutinative \\
             Bengali-Assamese Script \\
             Match Register of Examples \\ No Addtl. Explanations \\
             Output Fluent En \\
             Output Natural En \\
             Output Only Single Line of En \\
             SOV Word Order \\ 
             
        \end{tabular} \\
        \midrule
        \lus & Lus & \href{https://github.com/dhawan98/Gators_wmt26/blob/292f0922be4aea1aeedf9e2159f83dd52123c3af/translate_llm_manyshot_gemini.py#L151}{URL} & \begin{tabular}{l}
             Latin Script \\ 
             Match Example Register \\
             Match Example Style \\
             No Addtl. Explanations\\
             Output Only Single Line of Lus\\
             Output Std. Lus Orthography \\
             SOV Word Order \\ 
             Tonal w/o Changes in Script \\
             Tibeto Burman Family \\
             Verb-Final Agglutination \\
        \end{tabular}& \href{https://github.com/dhawan98/Gators_wmt26/blob/292f0922be4aea1aeedf9e2159f83dd52123c3af/translate_llm_manyshot_gemini.py#L165}{URL} & \begin{tabular}{l}
             No Addtl. Explanations\\
             Output Fluent En \\
             Output Natural En \\
             Output Only Single Line of En\\
        \end{tabular}\\
        \midrule
        \kha & Khasi & \href{https://github.com/dhawan98/Gators_wmt26/blob/292f0922be4aea1aeedf9e2159f83dd52123c3af/translate_llm_manyshot_gemini.py#L190}{URL} &\begin{tabular}{l}
             Austroasiatic Family \\
             Different Typology \\
             Gendered Prefix Agreement \\
             Latin Diacritics \\
             No Addtl. Explanations \\
             Output Only Single Line of Kha \\
             Polysynthetic Morph. \\
             Preserve Diacritics \\
             VOS Word Order \\
        \end{tabular} & \href{https://github.com/dhawan98/Gators_wmt26/blob/292f0922be4aea1aeedf9e2159f83dd52123c3af/translate_llm_manyshot_gemini.py#L205}{URL} & \begin{tabular}{l}
             Gendered Prefix Agreement \\
             No Addtl. Explanations\\
             Output Fluent En \\
             Output Natural En \\
             Output Only Single Line of En\\
             Polysynthetic Morph. \\
             VOS Word Order \\
        \end{tabular}\\ 
        \midrule
        \mni & Mni & \href{https://github.com/dhawan98/Gators_wmt26/blob/292f0922be4aea1aeedf9e2159f83dd52123c3af/translate_llm_manyshot_gemini.py#L229}{URL} & \begin{tabular}{l}
             Bengali Script \\
             Complex Verb Agglutination\\
             Match Example Style \\
             NFC Unicode Encoding \\
             No Addtl. Explanations \\
             No Bengali \\
             Output Only Single Line of Mni \\
             SOV Word Order \\
             Tibeto-Burman Family\\
             Unicode NFC Norm. \\
        \end{tabular}  & \href{https://github.com/dhawan98/Gators_wmt26/blob/292f0922be4aea1aeedf9e2159f83dd52123c3af/translate_llm_manyshot_gemini.py#L244}{URL} & \begin{tabular}{l}
            No Addtl. Explanations\\
            Output Fluent En\\
            Output Natural En\\
            Output Only Single Line of En\\
        \end{tabular} \\
        \midrule
        \mtei & Mni\_Mtei & \href{https://github.com/dhawan98/Gators_wmt26/blob/292f0922be4aea1aeedf9e2159f83dd52123c3af/translate_llm_manyshot_gemini.py#L268}{URL} & \begin{tabular}{l}
             Agglutinative\\
             Match Example Script\\
             Match Example Style\\
             Meitei Mayek Script\\
             No Addtl. Explanations\\
             No Bengali Script \\
             No Devanagari Script\\
             Output Only Single Line of Mtei\\
             SOV Word Order\\
             Tibeto-Burman Family\\
        \end{tabular} & \href{https://github.com/dhawan98/Gators_wmt26/blob/292f0922be4aea1aeedf9e2159f83dd52123c3af/translate_llm_manyshot_gemini.py#L281}{URL} & \begin{tabular}{l}
            No Addtl. Explanations\\
            Output Fluent En\\
            Output Natural En\\
            Output Only Single Line of En\\
             
        \end{tabular} \\
        \midrule
        \njz & Njz & \href{https://github.com/dhawan98/Gators_wmt26/blob/292f0922be4aea1aeedf9e2159f83dd52123c3af/translate_llm_manyshot_gemini.py#L304}{URL} & \begin{tabular}{l}
             Agglutinative\\
             Follow Example Orthography\\
             Latin Script\\
             No Addtl. Explanations\\
             Output Only Single Line of Njz\\
             Prefer Short Translations\\
             SOV Word Order\\
             Tibeto-Burman Family\\
             Very LoRes\\
        \end{tabular} & \href{https://github.com/dhawan98/Gators_wmt26/blob/292f0922be4aea1aeedf9e2159f83dd52123c3af/translate_llm_manyshot_gemini.py#L317}{URL} & \begin{tabular}{l}
            No Addtl. Explanations\\
            Output Fluent En\\
            Output Natural En\\
            Output Only Single Line of En\\
        \end{tabular} \\
        \bottomrule
    \end{tabular}
    \caption{We describe the prompting strategies we utilize for each Category~1 language in this table. Links in this table direct the reader to each prompt where it lives in our public Github repository.}
    \label{tab:prompts}
\end{table*}

\begin{table*}
    \centering
    \small
    \begin{tabular}{ll|c|p{4.5cm}|c|p{4.5cm}}
        \toprule
        \multicolumn{2}{c}{Language} &
        \multicolumn{2}{|c}{\texttt{En$\to$X}} & \multicolumn{2}{|c}{\texttt{X$\to$En}} \\
        \midrule
        Lang. & ISO & URL & Considerations & URL & Considerations \\
        \midrule
        \bodo & Bodo & \href{https://github.com/dhawan98/Gators_wmt26/blob/292f0922be4aea1aeedf9e2159f83dd52123c3af/translate_llm_manyshot_gemini.py#L343}{URL} & \begin{tabular}{l}
             Agglutinative Morph.\\
             Devanagari Script\\
             Examples Define Bodo\\
             Follow Examples Carefully\\
             Grammar Different From Hindi\\
             No Addtl. Explanations\\
             No Hindi, Sanskrit, or Marathi\\
             Not Indo-Aryan Family\\
             Output Only Single Line of Bodo\\
             SOV Word Order\\
             Tibeto-Burman Family\\
             Unicode NFC Normalization\\
        \end{tabular} & \href{https://github.com/dhawan98/Gators_wmt26/blob/292f0922be4aea1aeedf9e2159f83dd52123c3af/translate_llm_manyshot_gemini.py#L357}{URL} & \begin{tabular}{l}
             Devanagari Script\\
             No Addtl. Explanations\\
             Output Fluent En\\
             Output Natural En\\
             Output Only Single Line of En\\
             Tibeto-Burman Family\\ 
        \end{tabular} \\
        \midrule
        \trp & Trp & \href{https://github.com/dhawan98/Gators_wmt26/blob/292f0922be4aea1aeedf9e2159f83dd52123c3af/translate_llm_manyshot_gemini.py#L380}{URL} & \begin{tabular}{l}
             Agglutinative Morph.\\
             Follow Examples Closely\\
             Latin Script\\
             Match Example Orthography\\
             Match Example Script\\
             No Addtl. Explanations\\
             Output Only Single Line of Trp\\
             SOV Word Order\\
             Tibeto-Burman Family\\
             Very LoRes\\
        \end{tabular} & \href{https://github.com/dhawan98/Gators_wmt26/blob/292f0922be4aea1aeedf9e2159f83dd52123c3af/translate_llm_manyshot_gemini.py#L392}{URL} & \begin{tabular}{l}
             No Addtl. Explanations\\
             Output Fluent En\\
             Output Natural En\\
             Output Only Single Line of En\\
        \end{tabular} \\
        \midrule
        \mjw & Mjw & \href{https://github.com/dhawan98/Gators_wmt26/blob/292f0922be4aea1aeedf9e2159f83dd52123c3af/translate_llm_manyshot_gemini.py#L414}{URL} & \begin{tabular}{l}
             Agglutinative\\
             Extremely LoRes\\
             Latin Script\\
             Match Example Orthography\\
             No Addtl. Explanation\\
             Output Only Single Line of Mjw\\
             SOV Word Order\\
             Tibeto-Burman Family\\
             Tonal\\
        \end{tabular} & \href{https://github.com/dhawan98/Gators_wmt26/blob/292f0922be4aea1aeedf9e2159f83dd52123c3af/translate_llm_manyshot_gemini.py#L427}{URL} & \begin{tabular}{l}
             No Addtl. Explanations\\
             Output Fluent En\\
             Output Natural En\\
             Output Only Single Line of En\\
        \end{tabular} \\
        \midrule
        \nag & Nag & \href{https://github.com/dhawan98/Gators_wmt26/blob/292f0922be4aea1aeedf9e2159f83dd52123c3af/translate_llm_manyshot_gemini.py#L451}{URL} & \begin{tabular}{l}
             En and As/Bengali Vocabulary\\
             En-Based Creole\\
             Fewer Inflections than En Srcs\\
             Latin Script\\
             Match Example Orthography\\
             No Addtl. Explanations\\
             No En\\
             No SOV Word Order\\
             Output Only Nag Creole\\
             Output Single Line of Nag\\
             Simplified Morphology\\
             Simplified En-Based Phonology\\
             SVO Word Order\\
        \end{tabular} & \href{https://github.com/dhawan98/Gators_wmt26/blob/292f0922be4aea1aeedf9e2159f83dd52123c3af/translate_llm_manyshot_gemini.py#L466}{URL} & \begin{tabular}{l}
             No Addtl. Explanations\\
             Normalize Nag Form to Std. En\\
             Output Fluent En\\
             Output Natural En\\
             Output Single Line of Std. En\\
             Output Std. En Only\\
        \end{tabular} \\
        \midrule
        \tgj & Tgj & \href{https://github.com/dhawan98/Gators_wmt26/blob/292f0922be4aea1aeedf9e2159f83dd52123c3af/translate_llm_manyshot_gemini.py#L489}{URL} & \begin{tabular}{l}
             Agglutinative\\
             Extremely LoRes\\
             Latin Script\\
             Match Example Orthography\\
             Njz Relation\\
             No Addtl. Explanations\\
             Output Only Single Line of Tgj\\
             Shared Njz Vocab.\\
             SOV Word Order\\
             Tibeto-Burman Family\\
        \end{tabular} & \href{https://github.com/dhawan98/Gators_wmt26/blob/292f0922be4aea1aeedf9e2159f83dd52123c3af/translate_llm_manyshot_gemini.py#L502}{URL} & \begin{tabular}{l}
             No Addtl. Explanations\\
             Output Fluent En\\
             Output Natural En\\
             Output Only En\\
             Output Single Line of En\\ 
        \end{tabular} \\
        \bottomrule
    \end{tabular}
    \caption{We describe the prompting strategies we utilize for each Category~2 language in this table. Links in this table direct the reader to each prompt where it lives in our public Github repository.}
    \label{tab:prompts2}
\end{table*}

\end{document}